\PassOptionsToPackage{table,dvipsnames}{xcolor}
\documentclass[10pt,twocolumn,letterpaper]{article}

\usepackage[iccv]{iccv}      

\definecolor{iccvblue}{rgb}{0.21,0.49,0.74}
\usepackage[pagebackref,breaklinks,colorlinks,allcolors=iccvblue]{hyperref}

\def\paperID{10075} 
\def\confName{ICCV}
\def\confYear{2025}
\usepackage{times}  
\usepackage{graphicx} 
\usepackage{natbib}  
\usepackage{caption} 
\usepackage{algorithm}
\usepackage{algorithmic}

\usepackage{enumitem}
\usepackage{amssymb}
\usepackage{pifont}
\usepackage{xspace}
\usepackage{amsmath}
\usepackage{booktabs}       
\usepackage{colortbl} 
\usepackage{multirow}

\newcommand{\name}{TrackAny3D}
\newcommand{\nameo}{TrackAny3D~}

\usepackage{tikz}
\usepackage{amssymb}

\definecolor{lightgray}{RGB}{230,230,230}
\definecolor{Gray}{gray}{0.90}
\definecolor{LightCyan}{rgb}{0.82,0.82,1}
\usepackage{newfloat}
\usepackage{listings}
\title{TrackAny3D: Transferring Pretrained 3D Models for Category-unified\\ 3D Point Cloud Tracking}

\author{Mengmeng Wang$^{1,3}$\quad Haonan Wang$^{1}$\quad Yulong Li$^{1}$ \quad Xiangjie Kong$^{1,3}$ 
\\ \quad Jiaxin Du$^{1,3}$ \quad Guojiang Shen$^{1,3}$\thanks{Corresponding author: Guojiang Shen} \quad Feng Xia$^{2}$\\
$^{1}$ Zhejiang University of Technology \quad $^{2}$ RMIT University  \\ \quad $^{3}$ Zhejiang Key Laboratory of Visual Information Intelligent Processing\\
}

\begin{document}
\maketitle
\begin{abstract}
3D LiDAR-based single object tracking (SOT) relies on sparse and irregular point clouds, posing challenges from geometric variations in scale, motion patterns, and structural complexity across object categories. Current category-specific approaches achieve good accuracy but are impractical for real-world use, requiring separate models for each category and showing limited generalization.
To tackle these issues, we propose \name, the first framework to transfer large-scale pretrained 3D models for category-agnostic 3D SOT. We first integrate parameter-efficient adapters to bridge the gap between pretraining and tracking tasks while preserving geometric priors. Then, we introduce a Mixture-of-Geometry-Experts (MoGE) architecture that adaptively activates specialized subnetworks based on distinct geometric characteristics. Additionally, we design a temporal context optimization strategy that incorporates learnable temporal tokens and a dynamic mask weighting module to propagate historical information and mitigate temporal drift.
Experiments on three commonly-used benchmarks show that \nameo establishes new state-of-the-art performance on category-agnostic 3D SOT, demonstrating strong generalization and competitiveness. We hope this work will enlighten the community on the importance of unified models and further expand the use of large-scale pretrained models in this field.
\end{abstract}
\section{Introduction}
\label{sec:intro}
3D SOT on point clouds~\cite{sc3d,p2b,stnet} is a task of persistently localizing a target in a dynamic 3D scene. This task holds substantial potential for a wide range of applications, such as autonomous driving and mobile robotics. Unlike RGB-based tracking~\cite{siamcar,ostrack,citetracker}, which benefits from rich texture and color cues, 3D LiDAR-based SOT relies exclusively on sparse, irregular point clouds to infer a target’s 3D spatial pose. This geometric dependency introduces unique challenges: objects from different categories (e.g., cars, pedestrians) exhibit drastic variations in scale, motion patterns, and structural complexity. 
\begin{figure}
  \centering
  \includegraphics[width=1\linewidth]{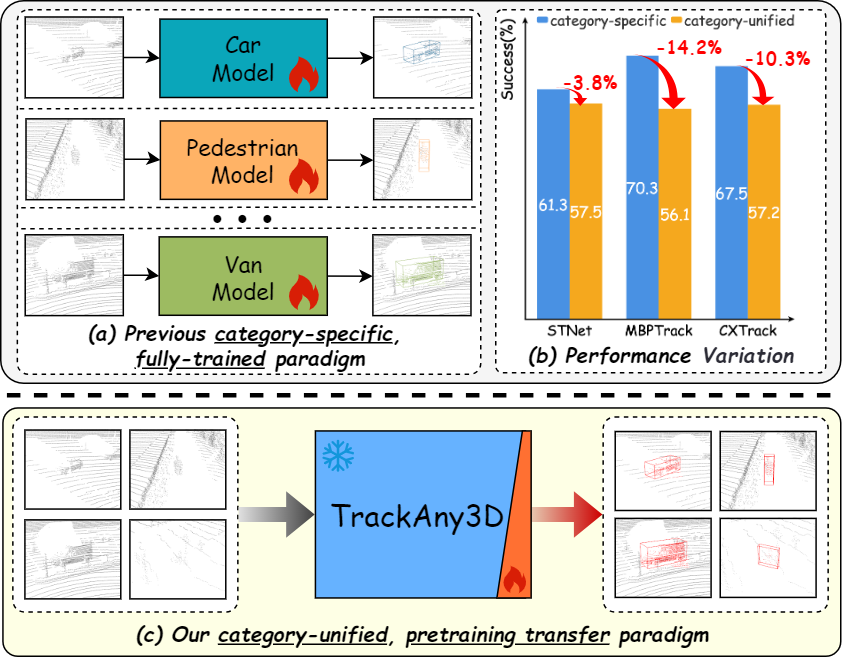} 
  \caption{\small \textbf{Comparison between Different Tracking Paradigms.} The previous category-specific, fully trained paradigm (a) employs multiple models, each learned for a specific category. We observed that they face significant performance drops when using cross-category training (on KITTI) (b). In contrast, our category-unified, pretraining transfer paradigm (c) uses a single shared model for all categories, which is efficiently transferred from pretrained 3D models.}\label{fig:intro}
  \vspace{-.5cm}
\end{figure}

To mitigate these challenges, existing methods~\cite{synctrack,mbptrack,p2p,streamtrack,cropnet} adopt a category-specific learning paradigm as shown in Fig.~\ref{fig:intro}(a), where plenty of dedicated models are independently trained and tested for each object type. While attaining leading accuracy, this paradigm is impractical for real-world deployment, as it demands prohibitive computational resources to train and store dozens of category-specific networks, and fails to generalize to novel categories, a critical limitation for open-world applications.
To validate the inter-class geometric variation challenge, we empirically observe that directly applying existing methods~\cite{mbptrack,stnet,cxtrack} to train unified models on all categories leads to severe performance degradation compared to their category-specific counterparts, as shown in Fig.~\ref{fig:intro}(b).
MoCUT~\cite{mocut} is the only method that attempts to solve this problem by explicitly encoding distinct attributes associated with different object categories. However, it mainly enforces uniformity through non-learnable constraints, which demands manual hyper-parameter tuning and limits generalization. Based on the above analysis, we are interested in the question: \textbf{How can we learn geometry-aware yet category-agnostic representations without introducing manual biases?}

The rise of large-scale pretrained models~\cite{clip,llama,pointmae,llava} provides a transformative and promising solution. In the fields of 2D vision \cite{st-adapter,action-clip,citetracker,m2clip} and natural language processing (NLP)~\cite{bert,roberta}, foundation models pretrained on web-scale data have demonstrated remarkable downstream generalization. This is enabled by parameter-efficient fine-tuning transfer (PEFT) techniques such as prompt tuning \cite{coop,vpt} or adapter modules \cite{adapter,clip-adapter}. Similarly, we believe that pretrained 3D point cloud models~\cite{pointclip,pointcontrast,pointmae} could provide valuable geometric priors for 3D SOT, potentially alleviating the aforementioned geometric disparity challenges to some extent.
However, extending this paradigm to 3D SOT remains largely unexplored and primarily faces three fundamental barriers: (i) distribution mismatch, where pretraining datasets (e.g., ShapeNet~\cite{shapenet}, ScanNet~\cite{scannet}) have limited category diversity and scene complexity compared to real-world tracking scenarios~\cite{kitti,Nuscenes}; (ii) persistent gaps, where pretrained models partially alleviate geometric disparities but fail to fully resolve the intrinsic conflict between geometric sensitivity; and (iii) lack of temporal modeling, as pretraining tasks focus on static shape reconstruction~\cite{pointclip,pointcontrast} or recognition~\cite{pointbert,pointmae}, while tracking requires modeling temporal coherence.

To address these challenges, we propose \name, the first framework to effectively transfer large-scale pretrained point cloud models for category-agnostic 3D SOT.
\nameo follows a novel category-unified, pretraining transfer paradigm, as shown in Fig.~\ref{fig:intro}(c), and consists of three core designs, each addressing one of the three aforementioned problems.
Specifically, we integrate a lightweight, two-path adapter into the Transformer layers.  One path within the adapter handles feature adaptation, while the other path modulates the intensity of this adaptation. This adapter dynamically aligns pretrained features with 3D SOT tasks while freezing the original pretrained network to preserve geometric priors and enhance learning efficiency. In order to further address the persistent gaps, we introduce Mixture-of-Geometry-Experts (MoGE) which consists of several expert subnetworks, where specialized experts are adaptively activated based on object geometry, resolving conflicts between divergent geometric patterns. Additionally, considering temporal modeling, we propose temporal context optimization strategies that propagate historical states through learnable temporal tokens and adaptively calibrate input information based on real-time geometric changes via a dynamic mask weighting mechanism.
Extensive experiments conducted on KITTI~\cite{kitti}, NuScenes~\cite{Nuscenes}, and Waymo~\cite{waymo} demonstrate that our \nameo achieves state-of-the-art (SOTA) performance under the category-unified setting, while also showcasing robust generalization capabilities.

Our main contributions can be summarized as: \textbf{1)}We propose \name, the first method to successfully adapt large-scale 3D pretrained models to open-world 3D SOT without category-specific tuning, by integrating lightweight adapters that enable effective knowledge transfer. \textbf{2)} We introduce a mixture-of-geometry-experts architecture where each expert subnetwork learns distinct geometric characteristics to ensure unified yet adaptive processing across diverse object categories. \textbf{3)} We design temporal propagation strategies with learnable temporal tokens coupled with a dynamic mask weighting mechanism, which jointly addresses temporal variations and state drift.

\section{Related Works}
\label{sec:related}
\subsection{3D Single Object Tracking}
Since the advent of the pioneering 3D LiDAR-based tracker SC3D~\cite{sc3d}, this field has witnessed rapid development in recent years~\cite{p2p,streamtrack,m2track,synctrack}. Current 3D SOT methods can be broadly categorized into three types. The first type is Siamese-based methods, which extract features from template and search regions using shared backbones and perform matching via similarity learning. Most existing tracking methods fall into this category~\cite{p2b,v2b,bat,votenet,sc3d,3dsiamrpn,stnet,cxtrack,mbptrack,osp2b,glt,p2p,scvtrack,m3sot}. 
The second type is motion-centric methods~\cite{m2track,voxeltrack,p2p,siammo,seqtrack3d}, which leverage the target's prior state from the previous frame to infer relative motion offsets in the current frame. 
Another type is unified modeling methods~\cite{synctrack,cropnet}, which jointly encode the template and search regions within a single Transformer architecture, unifying feature extraction and matching into an end-to-end framework. Our method adopts this approach to simplify the overall framework and maintain consistency with the input format of pretraining tasks.

Despite the progress made, most existing 3D SOT methods remain category-specific, meaning they are typically trained and fine-tuned for particular object categories, which is undesirable in real-world applications. Only recently has MoCUT~\cite{mocut} focused on this issue. However, MoCUT's solution often relies on manually designed rules and hyperparameter tuning. In this work, we address this problem by adapting the pretraining transfer paradigm and utilizing an adaptive network to overcome this limitation.

\subsection{PEFT in Point Clouds}
The rapid advancement of large-scale pretrained models, exemplified by CLIP\cite{clip}, LLaVA\cite{llava}, and Llama\cite{llama}, has driven the progress in core areas such as 2D vision and NLP through their superior representation learning capabilities. In the point cloud modality, several powerful large-scale models have also emerged, including PointCLIP\cite{pointclip}, Point-MAE\cite{pointmae}, Point-BERT\cite{pointbert}, CLIP2Point\cite{clip2point}, and RECON\cite{recon}, achieving excellent results in tasks like classification and generation.
The advancement of large-scale models has also propelled the progress of downstream tasks. Apart from full fine-tuning, PEFT techniques~\cite{lora,adapter,clip-adapter,tip-adapter,coop,vpt,maple} have been widely adopted, which allow models to be adapted efficiently with minimal resource consumption. 
In the point cloud field, research on PEFT remains relatively limited. Existing methods include IDAT\cite{idpt}, which combines DGCNN\cite{dgcnn} with instance-aware prompt extraction for geometric alignment; and DAPT\cite{dapt}, which integrates dynamic adapters with prompt tuning through internal prompts to adaptively capture domain variations.

However, in 3D SOT of the point cloud, the integration of large pretrained models with PEFT remains largely unexplored. To the best of our knowledge, the only related work is MemDisst~\cite{memdisst}, which uses 3D pretraining for initialization. Nevertheless, it lacks category-unified tracking, requires learning the entire network, and relies on distilling knowledge from a 2D pretrained tracker~\cite{ostrack} to ensure performance rather than leveraging PEFT techniques. This limits its efficiency and ability to fully exploit pretrained 3D knowledge. In contrast, this paper aims to fully utilize 3D pretrained models in combination with PEFT, addressing the limitations of category-specific approaches and significantly enhancing model generalizability and unification capabilities.

\begin{figure*}
  \centering
  \includegraphics[width=0.9\linewidth]{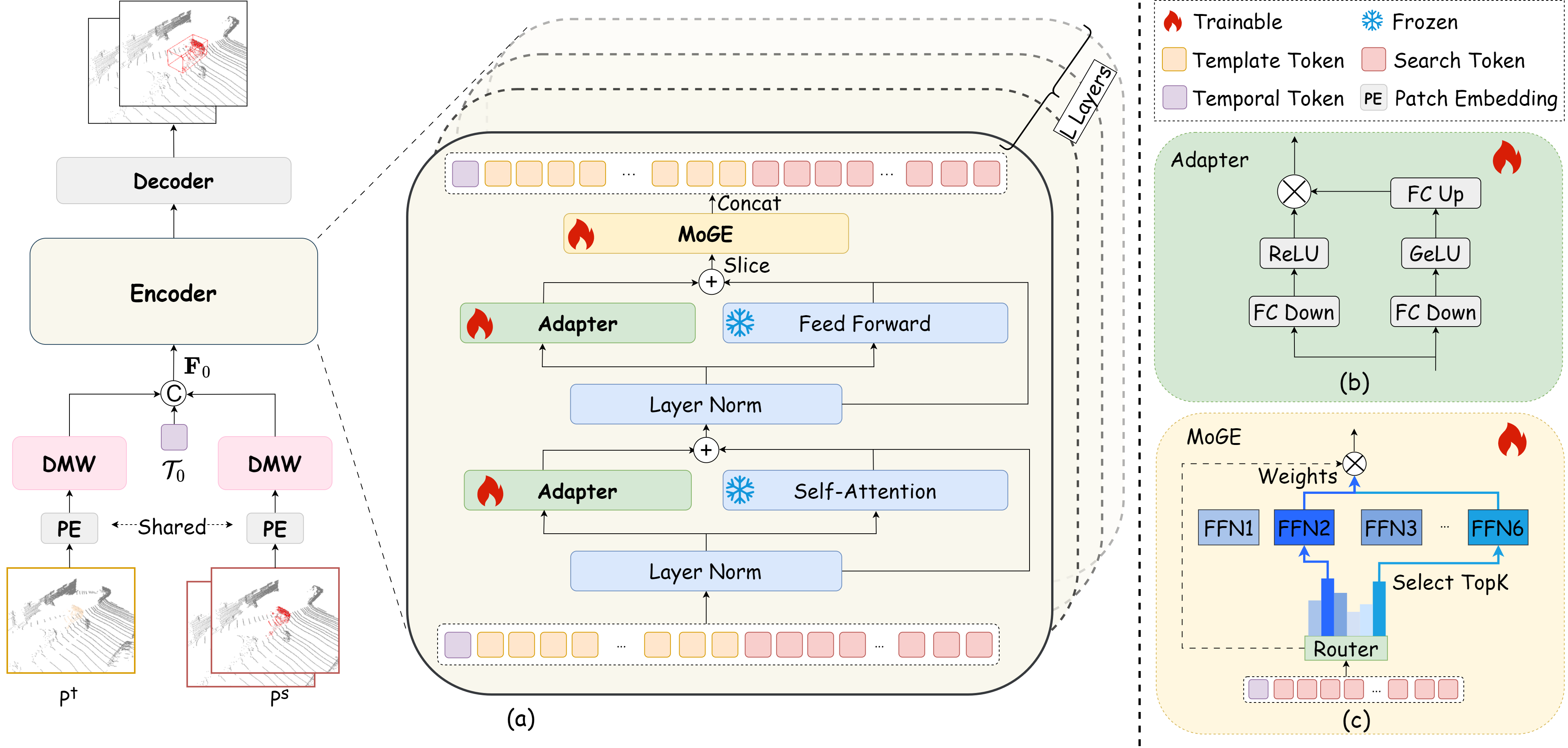} 
  \caption{\small \textbf{(a) An overview of our proposed \nameo architecture.} Our approach introduces a pretrained encoder, where we freeze the parameters of each Transformer layer. We then adapt it using a lightweight, two-path adapter and incorporate a mixture-of-geometry-expert (MoGE) module for further geometric modeling. Additionally, we introduce a learnable temporal token and a dynamic mask weight (DMW) mechanism to propagate and rectify temporal information. This ultimately results in an effective 3D tracker that shares a single model across all categories. \textbf{(b) Details of the adapter.} The adapter comprises two paths: one dedicated to performing the adaptation and the other to regulating the intensity of this adaptation. \textbf{(c) Details of MoGE.} It employs a Router to filter and select different geometry expert sub-networks (FFNs).}\label{fig:overview}
  \vspace{-.5cm}
\end{figure*}
\section{Methodology}
\subsection{Overview}
\noindent{\textbf{Task Definition.}}
In a dynamic 3D scene, 3D LiDAR-based SOT aims to localize a target within a sampled search region $\mathcal{P}^{s} = \{p_i^s\}_{i=1}^{N_s}$ in each frame, given a sampled template region $\mathcal{P}^{t} = \{p_i^t\}_{i=1}^{N_t}$, which corresponds to the target region from the initial or historical frames. The target's location is defined by a 3D bounding box (BBox) $B \in \mathbb{R}^7$, typically parameterized by the target's center coordinates $(x, y, z)$, orientation angle $\theta$ (the yaw angle around the vertical axis), and dimensions (width $w$, length $l$, height $h$). By applying translation and rotation to the template BBox $B_{t}$, the BBox $B_{s}$ for the current search frame can be calculated. Given that the size of the target remains consistent across all frames, only four parameters $(x, y, z, \theta)$ need to be predicted for $B_{t}$.

\noindent{\textbf{Framework Overview.}}
The overall framework is illustrated in Fig.~\ref{fig:overview}(a). For a given search region $\mathcal{P}^s$ and its corresponding template region $\mathcal{P}^t$, we first use a patch embedding layer to extract local information from each region and embed them into tokens. In addition to the point clouds themselves, we also introduce masks $\mathcal{M}^t$ and $\mathcal{M}^s$~\cite{m2track,mbptrack}. These masks are first processed through our dynamic mask weighting module (Sec.~\ref{sec:temporal}) and then added element-wise to their corresponding point cloud tokens. Furthermore, we incorporate a learnable temporal token $\mathcal{T}_0$ for temporal information propagation (Sec.~\ref{sec:temporal}). The template and search tokens, along with the temporal token, are concatenated to form the input $\mathbf{F}_0$ to the encoder.
The input $\mathbf{F}_0$ is then passed through an encoder transferred from a pretrained model, which consists of multiple frozen Transformer blocks augmented with our proposed parameter-efficient adapters (Sec.~\ref{sec:adapter}) and geometry-aware experts architecture (Sec.~\ref{sec:moe}). Finally, the output search tokens from the encoder are fed into a localization head~\cite{mbptrack} to predict bounding boxes for the input search region and the output encoded temporal token is retained to provide contextual cues for the next search frame.

\subsection{Efficient Pretrained Model Transfer}\label{sec:adapter}
\nameo utilizes RECON~\cite{recon} as the pretrained model, which is a powerful 3D representation learning framework combining the advantages of generative masked modeling and contrastive modeling. Here, we provide a brief introduction to its architecture. 
Formally, given an input point cloud $ \mathcal{P} \in \mathbb{R}^{N \times 3} $ with $N$ points, RECON employs a lightweight PointNet~\cite{pointnet} as the patch embedding layer to encode it into input embeddings $ \mathbf{F}_{0} \in \mathbb{R}^{N \times d} $, where $d$ represents the embedded feature dimension. Next, similarly to ViT~\cite{vit}, the encoder of RECON consists of $L$ Transformer layers, which encode the input tokens \( \mathbf{F}_{0} \). Specifically, each Transformer layer mainly consists of a standard multi-head self-attention (MHSA) block, layer normalizations (LN), and a feed-forward network (FFN). Formally, for the $i$-th layer:
\begin{align}
     \hat{\mathbf{F}}_{i-1} &= \text{MHSA}(\text{LN}(\mathbf{F}_{i-1})) + \mathbf{F}_{i-1},\\
    \mathbf{F}_i &= \text{FFN}(\text{LN}( \hat{\mathbf{F}}_{i-1})) +  \hat{\mathbf{F}}_{i-1},
\end{align}

To maintain consistency with the input of the pretrained model, we adopt a unified modeling framework that concatenates a learnable temporal token, the embedded point tokens of the template frame and the search frame to formulate $ \mathbf{F}_{0} \in \mathbb{R}^{(1+N_t+N_s) \times d} $, while performing feature extraction and matching through unified Transformer blocks.
In fact, the most straightforward method to transfer the pretrained model is to fully fine-tune it; however, we found this can lead to suboptimal performance (Table~\ref{tab:components}) and resource-intensive training. This is because when the model overwrites the knowledge learned during pretraining, it can result in a degradation of its original capabilities~\cite{st-adapter,action-clip}. Therefore, we explore the PEFT approach, which enables the model to adapt to new tasks with a small number of learnable parameters~\cite{dapt} while preserving the pretrained knowledge by keeping its core parameters frozen.

Specifically, as shown in Fig.~\ref{fig:overview}(b), our adapter module includes two paths: an adaptation path and a gated scoring path. The former contains a downsample projection layer with parameters $\mathbf{W}_{\text{dn}} \in \mathbb{R}^{d \times r}$, a GeLU activation function, and an upsample projection layer with parameters $\mathbf{W}_{\text{up}} \in \mathbb{R}^{r \times d}$. The gated scoring path, on the other hand, comprises a scoring weight matrix $\mathbf{W}_{s} \in \mathbb{R}^{d \times 1}$ and a ReLU activation function. This path is designed to compute a dynamic scaling factor for each token, which regulates the influence of the adaptation process in a data-driven manner. Then these two outputs are multiplied element-wise. Overall, for an input feature $\mathbf{F}_{i}$, the process of the adapter (AD) can be described as:
\begin{equation}
\text{AD}(\mathbf{F}_{i}) = \text{ReLU}\left(\mathbf{F}_{i}\mathbf{W}_{s} \right) \odot \text{GeLU}\left(\mathbf{F}_{i}\mathbf{W}_{dn}\right)\mathbf{W}_{up}
\end{equation}
This two-path design ensures that the adapter module can effectively control the contribution of the adapted features. We use two adapters added to each Transformer layer, parallel to the MHSA and FFN layers as:
\begin{align}
    \hat{\mathbf{F}}_{i-1} &= \text{MHSA}(\text{LN}(\mathbf{F}_{i-1})) + \mathbf{F}_{i-1}+\text{AD}(\text{LN}(\mathbf{F}_{i-1})),\\
    \mathbf{F}_i &= \text{FFN}(\text{LN}(\hat{\mathbf{F}}_{i-1})) + \hat{\mathbf{F}}_{i-1}+\text{AD}(\text{LN}(\hat{\mathbf{F}}_{i-1}))
\end{align}

\subsection{Mixture-of-Geometry-Experts}\label{sec:moe}
Although the above adapter enables efficient transfer learning without modifying the core parameters of the pretrained model, its performance still exhibits limitations in cross-category scenarios (Table~\ref{tab:components}). This is due to the fact that pretraining datasets come from different data domains (e.g., ShapeNet~\cite{shapenet} primarily consists of indoor objects), creating a significant gap with our real-world 3D SOT scenarios. Therefore, geometric disparities persist even though they are partially alleviated by leveraging geometric priors. Our solution draws inspiration from MoE~\cite{moe}, which learns multiple data bias views using a set of experts. Although MoE was originally proposed for building large pre-trained models, we adapt it to the context of 3D transfer learning for geometry-aware modeling, referred to as Mixture-of-Geometry-Experts (MoGE), and demonstrate its effectiveness in enhancing category-unified generalization.
\begin{figure}
  \centering
  \includegraphics[width=0.9\linewidth]{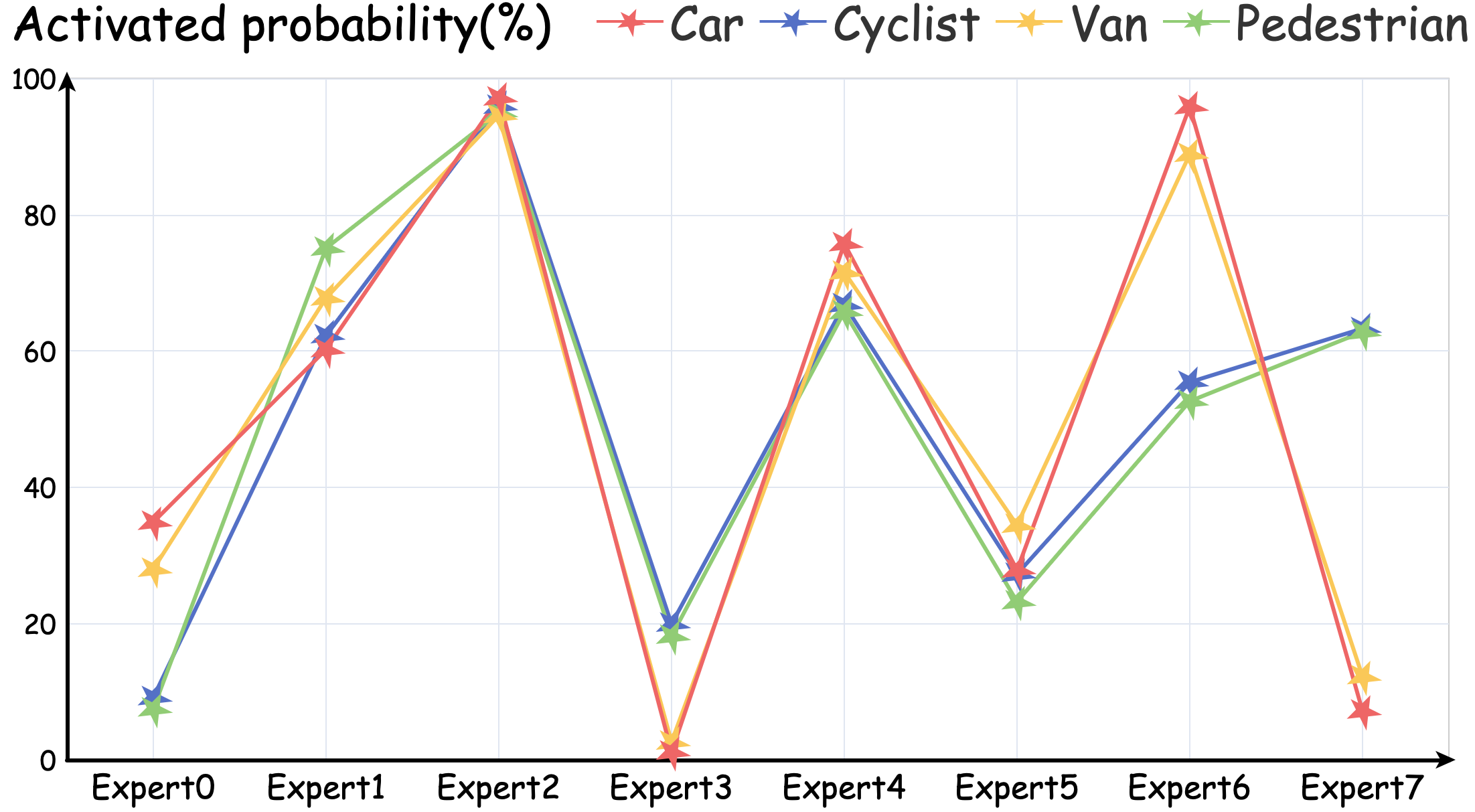} 
  \caption{\small \textbf{Geometry Sensitivity Analysis MoGE}. The distribution of activated experts for different object categories on KITTI. Pedestrian and Cyclist show significantly higher activation in Expert 3 and Expert 7, indicating that these experts excel at handling non-rigid and deformable geometry. In contrast, Van and Car exhibit higher activation in Expert 0 and Expert 6, suggesting that these experts focus on rigid structures.
Additionally, some experts (such as Expert 2 and Expert 4) display similar activation trends across all categories, implying that they may have learned general geometric features applicable to various object types.}\label{fig:moge}
  \vspace{-.3cm}
\end{figure}

The input $\textbf{Z}_j$ to the $j^{th}$ MoGE layer is constructed by concatenating the temporal token with the search tokens. As shown in Fig.~\ref{fig:overview}(c), the MoGE layer consists of $M$ geometry experts $\{\{\mathbf{E}_{j}^{m}\}_{j = 1}^{L}\}_{m = 1}^{M}$, where $\mathbf{E}_{j}^{m}$ represents the $m^{th}$ expert at the $j^{th}$ layer and has a same structure as the FFN. The routing algorithm of MoGE determines which experts process inputs. Here we employ the Top-K gating routing mechanisms as our Router $\{\mathbf{R}_{j}\}_{j = 1}^{L}$ to make decisions using a learnable gate network, which includes an expert embedding $\textbf{W}_{j}^R \in \mathbb{R}^{d \times M}$ to transfer features to scores.
Specifically, the MoGE output can be represented as:
\begin{align}
&\mathbf{E}_{j}(\textbf{Z}_j) = \sum_{m=1}^{M} \mathbf{R}_{j}(\textbf{Z}_j, K) \mathbf{E}_{j}^{m}(\textbf{Z}_j), \\
&\mathbf{R}_{j}(\textbf{Z}_j, K) = \text{Softmax}\left(\text{Top-K}\left(\textbf{Z}_j\textbf{W}_{j}^R, K \right) \right)
\end{align}
Only $K$ ($K < M$) experts will be activated, receiving the input and performing adaptive fusion before outputting the result. As shown in the Fig.~\ref{fig:moge}, MoGE demonstrates its design effectiveness by performing adaptive selection based on geometric characteristics rather than relying solely on class labels. We place our MoGE layer after the FFN inside Transformer blocks to avoid disrupting the original structure of the pretrained model.

\begin{table*}
\caption{Comparison with SOTA methods on KITTI. \colorbox{green!10}{\quad} and \colorbox{red!10}{\quad} refer to category-specific methods and category-unified methods, respectively. \textbf{Bold} and \underline{underline} denote current and previous best performance.``TP'' denotes Tunable Parameters.}
\label{tab:kitti}
\begin{center}
\vspace{-0.3cm}
    \resizebox{0.9\textwidth}{!}{
    \small
    \begin{tabular}{c|c|c|cccc|c}
          \toprule[0.4mm]
          \rowcolor{black!5} Method & FPS&TP(M) &Car [6,424] & Pedestrian [6,088] &  Van [1,248] & Cyclist [308] & Mean [14,068] \\
          \midrule
          \rowcolor{green!5} P2B \citep{p2b}& 40 &1.34& 56.2 / 72.8   & 28.7 / 49.6 & 40.8 / 48.4 & 32.1 / 44.7 & 42.4 / 60.0 \\
          \rowcolor{green!5} V2B \citep{v2b}&37&1.36 &70.5 / 81.3   & 48.3 / 73.5 & 50.1 / 58.0 & 40.8 / 49.7 & 58.4 / 75.2 \\
          \rowcolor{green!5} PTTR \citep{pttr}& 50&2.27&65.2 / 77.4   & 50.9 / 81.6 & 52.5 / 61.8 & 65.1 / 90.5 & 57.9 / 78.2 \\
          \rowcolor{green!5} STNet \citep{stnet}& 35&1.66&72.1 / 84.0& 49.9 / 77.2& 58.0 / 70.6 &73.5 / 93.7 &61.3 / 80.1\\
          \rowcolor{green!5} GLT-T \citep{glt}&30&2.60 &68.2 / 82.1   & 52.4 / 78.8 & 52.6 / 62.9 & 68.9 / 92.1 & 60.1 / 79.3 \\
          \rowcolor{green!5} CXTrack \citep{cxtrack}& 29&18.30&69.1 / 81.6 &67.0 / 91.5 &60.0 / 71.8 &74.2 / \underline{94.3} &67.5 / 85.3\\
           \rowcolor{green!5} M$^2$Track \citep{m2track} & 57&2.24&65.5 / 80.8   & 61.5 / 88.2 & 53.8 / 70.7 & 73.2 / 93.5 & 62.9 / 83.4 \\
          \rowcolor{green!5} SyncTrack \citep{synctrack}& 45&1.47&73.3 / \underline{85.0} &54.7 / 80.5 &60.3 / 70.0& 73.1 / 93.8 &64.1 / 81.9\\
           \rowcolor{green!5} MBPTrack \citep{mbptrack}& 50&7.38&\underline{73.4} / 84.8 &\underline{68.6} / \underline{93.9} &\textbf{61.3} / 72.7 &\underline{76.7} / \underline{94.3} &\underline{70.3} / 87.9\\
            \rowcolor{green!5} PTTR++ \citep{pttr++}& 43 &-&\underline{73.4} / 84.5 &55.2 / 84.7 &55.1 / 62.2 &71.6 / 92.8 &63.9 / 82.8\\
             \rowcolor{green!5} M3SOT \citep{m3sot}& 38 &-&\textbf{75.9 / 87.4} &66.6 / 92.5 &59.4 / \underline{74.7} &70.3 / 93.4 &\underline{70.3} / \textbf{88.6}\\
               \rowcolor{green!5} StreamTrack \citep{streamtrack}& 41&-&72.6 / 83.7 &\textbf{70.5} / \textbf{94.7} &\underline{61.0 }/ \textbf{76.9} &\textbf{78.1} / \textbf{94.6} &\textbf{70.8} / \underline{88.1}\\
           \midrule
           \rowcolor{red!5} STNet \citep{stnet}& 35&1.66&\underline{69.5} / \underline{80.5} &43.8 / 67.4 &58.1 / 68.2 & 73.8 / \underline{94.0} &57.5 / 74.0\\
            \rowcolor{red!5} CXTrack \citep{cxtrack}& 29&18.30&60.2 / 72.6 &54.6 / 81.6 &57.6 / 70.0 & 44.4 / 57.0 &57.2 / 75.9\\
            \rowcolor{red!5} M$^2$Track \citep{m2track}& 57&2.24&62.9 / 77.0 &57.4 / 84.4 &66.2 / \underline{80.7}& \underline{76.0} / 93.7 &61.1 / 80.9\\
          \rowcolor{red!5} MBPTrack \citep{mbptrack}& 50&7.38&62.3 / 72.1 &50.2 / 80.9 &\underline{66.6} / 78.2 &71.8 / 92.2 &56.1 / 74.9\\
          \rowcolor{red!5} SiamCUT \citep{mocut} & 36&2.06& 58.1 / 73.9 & 48.2 / 76.2 & 63.1 / 74.9 & 36.7 / 47.4  & 54.0 / 74.6 \\
          \rowcolor{red!5} MoCUT \citep{mocut}& 48&2.34&67.6 / \underline{80.5} & \textbf{63.3} / \textbf{90.0} & 64.5 / 78.8  & \textbf{76.7} / \textbf{94.2}  & \underline{65.8} / \underline{85.0} \\ \hline
          \rowcolor{red!5} \nameo&28 &5.30&\textbf{73.4} / \textbf{85.2} & \underline{59.6} / \underline{85.6} & \textbf{70.0} / \textbf{82.8}  & 74.7 / \underline{94.0}  & \textbf{67.1} / \textbf{85.4} \\
          \bottomrule[0.4mm]
    \end{tabular}}
\end{center}
\vspace{-0.7cm}
\end{table*}
\subsection{Temporal Context Optimization}\label{sec:temporal}
The initial pretrained model learns representations for static tasks, whereas tracking is inherently a dynamic task. Therefore, we explore additional temporal modeling methods. Inspired by prompt learning~\cite{coop,maple,odtrack,vitaclip}, we first define a learnable initial temporal token $\mathcal{T}_0\in \mathbb{R}^{1\times d}$ that aims to sufficiently interact with all template and search tokens throughout the encoder. In this way, the temporal token absorbs spatiotemporal representations relevant to the current time step.

\begin{figure}
  \centering
  \includegraphics[width=0.95\linewidth]{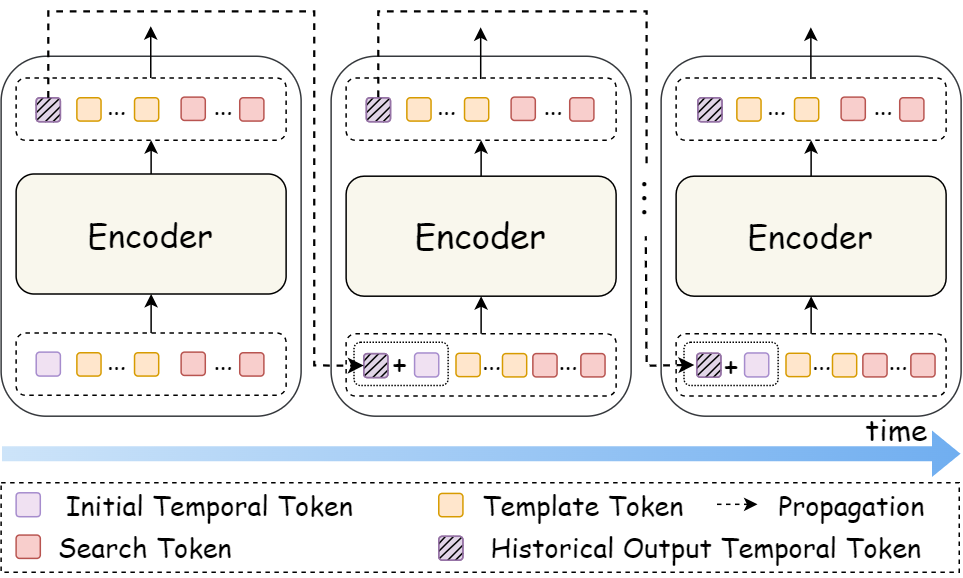} 
  \caption{\small \textbf{Temporal token  propagation}. The temporal token captures key features across different frames and propagates to subsequent frames sequentially. }\label{fig:token}
  \vspace{-.3cm}
\end{figure}

In detail, for a sequence of $ T $ frames, $\mathcal{T}_0$ will be propagated and updated along the time dimension, as illustrated in Fig.~\ref{fig:token}. 
At time step $ t $, the input temporal token is updated to $\mathcal{T}_0^t$, which is obtained by combining the learned initial temporal token $\mathcal{T}_0$ with the historical output temporal token $\mathcal{T}_{out}^{t-1}$ from the most recent historical frame's encoder output:
\begin{equation}
\mathcal{T}_0^t = \mathcal{T}_0 + \mathcal{T}_{out}^{t-1}
\end{equation}
Note that if $ t = 1 $, then $\mathcal{T}_0^t = \mathcal{T}_0$. The propagated token $\mathcal{T}_0^t$ is then integrated with current template and search tokens via the encoder. For the training process, $\mathcal{T}_0$ is randomly initialized and continuously updated along with the network. During testing, $\mathcal{T}_0$ is loaded with the trained values. This operation preserves temporal coherence by propagating historical features while avoiding complex computations. 

In addition, to address spatiotemporal variations across object categories, we propose a dynamic mask weighting (DMW) mechanism. We first construct masks~\cite{m2track} for the input point clouds, defining a target-oriented mask $\mathcal{M}^t$ where object regions are assigned a value of 0.8 and background regions a value of 0.2, while the search frame uses a uniform mask $\mathcal{M}^s$ initialized to 0.5. We then propose to apply learnable weights $\mathbf{\beta}^t \in \mathbb{R}^{N_t \times 1}$, $\mathbf{\beta}^s \in \mathbb{R}^{N_s \times 1}$ for template and search regions, which adaptively scale the masks through element-wise multiplication. The adjusted masks are then simply added to the embedded input tokens. Then, the whole process of the dynamic mask weighting module can be expressed as:
\begin{align}
\tilde{\mathcal{F}}^* = \mathcal{F}^* + \mathcal{M}^* \odot \mathbf{\beta}^*, \quad \forall * \in \{t, s\}
\end{align}
By jointly optimizing $\mathbf{\beta}^*$ during training, our method not only enhances mask quality through adaptive temporal noise suppression but also dynamically adjusts mask emphasis based on category-specific characteristics without manual hyperparameter tuning.

\section{Experiments}
\subsection{Implementation Details.}
We conduct comprehensive experiments using three widely-used datasets: KITTI~\cite{kitti}, NuScenes~\cite{Nuscenes}, and Waymo Open Dataset (WOD)~\cite{waymo}. We sample clips from each sequence to form training samples. The length of each clip is set to 3 frames. Specifically, for the template and search regions, we set the number of input points as $N_t$=128 and $N_s$=128 after farthest point sampling, respectively. Our model is built on the pretrained RECON~\cite{recon} with its original parameters frozen. The bottleneck dimension $r$ of our adapters is set to 72. For MoGE module, we use $M$=8 experts and select $K$=4. The FFN for MoGE, the intermediate bottleneck dimension is set to 1/8 of the Transformer dimension. The MoGE layer is placed at every even-number layer. Inference for our model was conducted on a single NVIDIA RTX3090 GPU.\\
\subsection{Comparison with State-of-the-art Methods}
\begin{table*}
\caption{Comparison with SOTA  on NuScenes.}
\vspace{-0.3cm}
\label{tab:nuscenes}
\begin{center}
    \resizebox{0.95\textwidth}{!}{
    \small
    \begin{tabular}{c|ccccc|c}
          \toprule[0.4mm]
         \rowcolor{black!5} Method&Car [64,159]&Pedestrian [33,227]&Truck [13,587]&Trailer [3,352]&Bus [2,953]&Mean [117,278] \\
          \midrule
          \rowcolor{green!5} SC3D \citep{sc3d}& 22.31 / 21.93   & 11.29 / 12.65 & 30.67/ 27.73 & 35.28 / 28.12 & 29.35 / 24.08 & 20.70 / 22.20 \\
          \rowcolor{green!5} P2B \citep{p2b}& 38.81 / 43.18  & 28.39 / 52.24 & 42.95 / 41.59 & 48.96 / 40.05 & 32.95 / 27.41 & 36.48 / 45.08\\
          \rowcolor{green!5} PTT \citep{ptt}& 41.22 / 45.26   & 19.33 / 32.03 & 50.23 / 48.56& 51.70 / 46.50 & 39.40 / 36.70 & 36.33 / 41.72 \\
          \rowcolor{green!5} PTTR \citep{pttr}& 51.89 / 58.61   & 29.90 / 45.09 & 45.30 / 44.74 & 45.87 / 38.36 & 43.14 / 37.74 & 44.50 / 52.07 \\
          \rowcolor{green!5} GLT-T \citep{glt}& 48.52 / 54.29 & 31.74 / 56.49 & 52.74 / 51.43 & 57.60 / 52.01 & 44.55 / 40.69 & 44.42 / 54.33 \\
          \rowcolor{green!5} CXTrack \citep{cxtrack}&  42.64 / 47.96  &  32.75 / 59.33  &  40.31 / 38.73  & 51.48 / 41.36 & 38.76 / 30.67  &  39.72 / 49.51 \\
          \rowcolor{green!5} M$^2$Track \citep{m2track}& 55.85 / 65.09  & 32.10 / 60.72 & 57.36 / 59.54 & 57.61 / 58.26 & 51.39 / 51.44& 49.32 / 62.73 \\
           \rowcolor{green!5} SeqTrack3D\cite{seqtrack3d}& \textbf{62.55} / \textbf{71.46} & \underline{39.94} / 68.57 & 60.97 / 63.04 & \textbf{68.37} / \underline{61.76} & 54.33 / 53.52 & \underline{55.92} / 68.94 \\
           \rowcolor{green!5} MBPTrack\cite{mbptrack}& \underline{62.47} / 70.41 & \textbf{45.32} / \textbf{74.03} & \underline{62.18} / \underline{63.31} & 65.14 / 61.33 & \underline{55.41} / 51.76 & \textbf{57.48} / \textbf{69.88} \\
           \rowcolor{green!5} PTTR++\cite{pttr++}& 59.96 / 66.73 & 32.49 / 50.50 & 59.85 / 61.20 & 54.51 / 50.28 & 53.98 / 51.22 & 51.86 / 60.63 \\
           \rowcolor{green!5} SCVTrack\cite{scvtrack}& 58.90 / 67.70 & 34.50 / 61.50 & 60.60 / 61.40 & 59.50 / 60.10 & 54.30 / \underline{53.60} & 52.10 / 64.70 \\
           \rowcolor{green!5} StreamTrack\cite{streamtrack}& 62.05 / \underline{70.81} & 38.43 / \underline{68.58 }& \textbf{64.67} / \textbf{66.60 }& \underline{66.67} / \textbf{64.27} & \textbf{60.66} / \textbf{59.74} & 55.75 / \underline{69.22} \\
          \midrule
           \rowcolor{red!5} CXTrack \citep{cxtrack}&  43.43 / 47.42  &  34.19 / 61.53  & 48.24 / 44.84  & 54.52 / 42.41  & 43.21 / 35.41  &  41.69 / 50.67  \\
            \rowcolor{red!5}  M$^2$Track \citep{m2track}& 50.76 / 57.62  &  28.35 / 53.53  & 55.97 / 57.39  & 47.87 / 42.09  &  51.65 / 49.10  & 44.98 / 55.76  \\
            \rowcolor{red!5}  MBPTrack \citep{mbptrack}& 55.77 / 61.46  &  \underline{37.62 } / \underline{65.38}  &  58.99 / 57.23  &  58.11 / 46.34  &  \underline{58.37} / 54.20  &  51.13 / 61.47  \\
          \rowcolor{red!5} SiamCUT \citep{mocut}&40.96 / 44.91& 31.42 / 53.80&53.91 / 52.65&\underline{63.29} / 58.21&41.03 / 38.01&40.41 / 48.54 \\
          \rowcolor{red!5} MoCUT \citep{mocut}& \underline{57.32} / \underline{66.01} & 33.47 / 63.12 & \underline{61.75} / \textbf{64.38} &60.90 / \textbf{61.84} &57.39 / \underline{56.07}&\underline{51.19} / \underline{64.63} \\ \hline
          \rowcolor{red!5} \nameo& \textbf{59.30} / \textbf{66.46} & \textbf{40.37} / \textbf{68.70} & \textbf{62.70} / \underline{62.80} &\textbf{66.12} / \underline{59.20} &\textbf{61.01} / \textbf{58.02}&\textbf{54.57} / \textbf{66.25} \\
          \bottomrule[0.4mm]
    \end{tabular}}
\end{center}
\vspace{-0.6cm}
\end{table*}
In this section, we conduct extensive comparisons on three datasets, categorizing all methods into two experimental settings: \colorbox{green!10}{category-specific} and \colorbox{red!10}{category-unified}. In the category-unified setting, to facilitate fair comparisons, we reproduce the results of several SOTA methods, including STNet~\cite{stnet}, M$^2$Track~\cite{m2track}, CXTrack~\cite{cxtrack}, and MBPTrack~\cite{mbptrack}, using their officially open-sourced code.

\noindent{\textbf{Results on KITTI.}} In Table~\ref{tab:kitti}, we compare our method against SOTA methods on KITTI. Recent category-specific models such as STNet, MBPTrack, and StreamTrack have shown steady improvements in accuracy, indicating rapid advancements in single-category model design. However, we observe that when these models are trained under the category-unified setting, their performance drops significantly. For instance, STNet, CXTrack, and MBPTrack experience performance declines of 3.8\%, 10.3\%, and 14.2\%, respectively.
In contrast, our method is comparable to those of category-specific methods and achieves the best results when trained on all categories together, reaching a success rate of 67.1\%, surpassing all other methods in this setting, including the latest MoCUT, which focuses on the same problem. Specifically, our method outperforms MoCUT by 1.3\% overall and by 6.0\% in the Car category. 
These results demonstrate the effectiveness of our approach, showcasing its ability to generalize well across multiple categories while maintaining high performance.\\
\noindent{\textbf{Results on NuScenes}}. Table~\ref{tab:nuscenes} presents the comparison results on NuScenes. Our method achieves a consistent and significant performance gain compared to previous SOTA methods, such as MoCUT, under the category-unified setting, while remaining competitive compared to category-specific methods. Additionally, we observe that in this dataset, due to the large amount of data for common categories like Cars and Pedestrians, the performance of less frequent categories (including Trailers, Trucks, Buses) is significantly improved, even surpassing methods like MBPTrack, which are specifically trained for certain categories.
Similar to the trend observed on KITTI, competing methods still exhibit a substantial performance drop between their category-unified and category-specific results; for example, MBPTrack shows a noticeable degradation in performance of 6.35\%. In contrast, our \nameo achieves the best results under the category-unified setting, with results on individual classes such as Bus even surpassing all single-category learning models. This highlights the superiority of our method in addressing geometric disparity issues.\\
\textbf{Results on WOD.} In Table~\ref{tab:waymo}, we perform direct inference on the Vehicle of WOD using our model trained on KITTI, thereby validating the generalization capability of our approach. For the category-specific methods, following their work, we use their KITTI Car models. The results indicate that \nameo demonstrates strong competitiveness and achieves the best tracking performance with 64\%, outperforming all other methods, including those designed for category-specific tasks. Specifically, compared to MoCUT, our method improves performance by 2.1\%, and compared to MBPTrack under the category-unified setting, we achieve a lead of 5.8\%. These results highlight that \name, under the new paradigm of pretrained model transfer, possesses excellent generalization capabilities, effectively addressing cross-dataset and cross-category challenges.

\begin{table}[tbp]
  \centering
  \vspace{-0.3cm}
  \caption{Comparison with SOTA on WOD.}
  \small
    \scalebox{1}{
     \setlength{\tabcolsep}{2.2pt}
    \begin{tabular}{c|ccc|c}
    \toprule
    \multirow{2}[4]{*}{Method} & \multicolumn{4}{c}{Vehicle[185731]} \\
\cmidrule{2-5}          & Easy & Medium & Hard & Mean \\
    \midrule
     \rowcolor{green!5}P2B\citep{p2b}   & 57.1 / 65.4 & 52.0 / 60.7 & 47.9 / 58.5 & 52.6 / 61.7 \\
     \rowcolor{green!5}BAT\citep{bat}   & 61.0 / 68.3 & 53.3 / 60.9 & 48.9 / 57.8 & 54.7 / 62.7 \\
     \rowcolor{green!5}V2B\cite{v2b}   & 64.5 / 71.5 & 55.1 / 63.2 & 52.0 / 62.0 & 57.6 / 65.9 \\
     \rowcolor{green!5}STNet\citep{stnet} & 65.9 / 72.7 & 57.5 / 66.0 & 54.6 / 64.7 & 59.7 / 68.0 \\
     \rowcolor{green!5}TAT\citep{tat}   & 66.0 / 72.6 & 56.6 / 64.2 & 52.9 / 62.5 & 58.9 / 66.7 \\
     \rowcolor{green!5}CXTrack\citep{cxtrack} & 63.9 / 71.1 & 54.2 / 62.7 & 52.1 / 63.7 & 57.1 / 66.1 \\
     \rowcolor{green!5}M$^2$Track\citep{m2track} & \underline{68.1} / \underline{75.3} & \textbf{58.6} / \underline{66.6} & \underline{55.4} / \underline{64.9} & 61.1 / 69.3 \\
     \rowcolor{green!5}MBPTrack\citep{mbptrack} & \textbf{68.5} / \textbf{77.1} & \underline{58.4} / \textbf{68.1} & \textbf{57.6} / \textbf{69.7} & \textbf{61.9} / \textbf{71.9} \\
     \rowcolor{green!5}SiamDisst\citep{memdisst} & - & - & - & \underline{61.2} / \underline{70.5} \\
     \rowcolor{green!5}MemDisst\citep{memdisst} & - & - & - & \textbf{61.9} / \textbf{71.9} \\
     \midrule
     \rowcolor{red!5}CXTrack\citep{cxtrack} &60.7 / 67.6 & 50.5 / 57.7 & 46.0 / 55.8 & 52.8 / 60.7 \\
      \rowcolor{red!5}STNet\citep{stnet} &67.5 / \underline{75.2} & 59.2 / 68.2 & 55.6 / \underline{66.4} & 61.1 / \underline{70.2} \\
     \rowcolor{red!5}MBPTrack\citep{mbptrack} &65.6 / 74.4 &55.4 / 64.6 &52.5 / 63.2 &58.2 / 67.7 \\
     \rowcolor{red!5}SiamCUT\citep{mocut} & 58.3 / 66.0 & 50.8 / 60.8 & 49.2 / 59.1 & 53.0 / 62.2 \\
     \rowcolor{red!5}MoCUT\citep{mocut} & \underline{68.3} / 75.0 & \underline{59.4} / \underline{66.9} & \underline{57.1} / 66.3 & \underline{61.9 }/ 69.7 \\ \hline
    \rowcolor{red!5}\nameo & \textbf{72.6} / \textbf{80.2} & \textbf{60.5} / \textbf{69.6} & \textbf{57.5} / \textbf{68.9} & \textbf{64.0} / \textbf{73.3} \\
    \bottomrule
    \end{tabular}%
    }
  \vspace{-0.6cm}
  \label{tab:waymo}%

\end{table}%
\subsection{Ablation Studies}

\begin{table*}
\small
\caption{Ablations of different model components. ``\colorbox{lightgray}{\quad}"  represent the final setting of our \name. ``FF" denotes Full Fine-tuning. ``AD" denotes Adapter Tuning. ``TT" denotes Temporal Token.}
\vspace{-0.3cm}
\label{tab:components}
\begin{center}
    \resizebox{0.8\textwidth}{!}{
    \small
    \begin{tabular}{ccccc|ccccc}
      \toprule[0.3mm]
        FF  & AD &MoGE & TT  & DMW  & Car & Pedestrian & Van & Cyclist & Mean \\

      \midrule
     \ding{51} & \ding{55} & \ding{55} & \ding{55} & \ding{55} & 69.8 / 83.6 & 53.6 / 81.6 & 62.2 / 73.6 & 65.8 / 90.8 & 62.0 / 82.0 \\
     \ding{51} & \ding{55} & \ding{51} & \ding{51} & \ding{51} & 72.5 / 84.0 & 57.9 / 82.0 & 70.8 / 82.9 & 72.8 / 93.7 & 66.0 / 83.3 \\ \hline
      \ding{55} & \ding{51} & \ding{55} & \ding{55} & \ding{55} & 71.8 / 85.6 & 55.8 / 83.9 & 69.8 / 82.7 & 74.1 / 94.1 & 64.8 / 84.8 \\
      \ding{55} & \ding{51} & \ding{51} & \ding{55} & \ding{55} & 72.5 / 84.8 & 56.3 / 83.5 & 71.2 / 83.2 & 67.4 / 92.1 & 65.3 / 84.3 \\
     \ding{55} & \ding{51} & \ding{51} & \ding{51} & \ding{55} & 72.5 / 84.4 & 58.1 / 85.7 & 71.9 / 83.8 & 72.4 / 89.7 & 66.2 / 85.0 \\
      \rowcolor{lightgray}\ding{55} & \ding{51} & \ding{51} & \ding{51} & \ding{51} & 73.4 / 85.2 & 59.6 / 85.6 & 70.0 / 82.8 & 74.7 / 94.0 & 67.1 / 85.4 \\
      \bottomrule[0.4mm] 
    \end{tabular}}
\end{center}
\vspace{-0.4cm}
\end{table*}
\begin{table}[htbp]
  \centering
  \small
 \vspace{-0.3cm}
  \caption{Ablations for positions of adapters(AD) and MoGE.}
  \setlength{\tabcolsep}{1.1pt}
   \scalebox{0.98}{
    \begin{tabular}{c|c|ccccc}
    \toprule
     & Position & Car & Pedestrian & Van & Cyclist & Mean \\
    \hline
   \multirow{3}{*}{\rotatebox[origin=c]{90}{AD}} &  \cellcolor{lightgray} all layers & \cellcolor{lightgray}73.4/85.2& \cellcolor{lightgray} 59.6/85.6 &  \cellcolor{lightgray}70.0/82.8 &  \cellcolor{lightgray}74.7/94.0 & \cellcolor{lightgray} 67.1/85.4 \\
                             & even layers & 69.0/80.0 & 60.0/86.0 & 72.3/84.4 & 73.4/93.1 & 65.5/83.3 \\
                             & last layer & 70.5/80.7 & 51.2/73.9 & 70.3/81.5 & 71.5/92.9 & 62.2/78.1 \\
    \cmidrule{1-7}
    \multirow{3}{*}{\rotatebox[origin=c]{90}{MoGE}}   & all layers & 71.8/83.5 & 53.6/81.7 & 69.5/81.3 & 72.5/93.5 & 63.7/82.7 \\
                             &  \cellcolor{lightgray}even layers & \cellcolor{lightgray}73.4/85.2 & \cellcolor{lightgray} 59.6/85.6 & \cellcolor{lightgray} 70.0/82.8 & \cellcolor{lightgray} 74.7/94.0 & \cellcolor{lightgray} 67.1/85.4 \\
                             & last layer & 70.2/81.8 & 50.3/78.3 & 71.5/82.8 & 71.7/92.6 & 61.7/80.6 \\
    \bottomrule
    \end{tabular}%
    }
  \label{tab:position}%
\end{table}%

\begin{table}[htbp]
  \centering
   \vspace{-0.2cm}
  \caption{Ablations for length of temporal propagation.}
   \small
   \setlength{\tabcolsep}{3pt}
   \scalebox{0.98}{
       \begin{tabular}{c  |ccccc}
    \toprule
    Length &Car &Pedestrain &Van &Cyclist & Mean  \\ 
\hline
  2&  69.5/80.9 &  56.1/82.5 &73.0/84.5 &66.9/90.8 & 64.0/82.1 \\
  \cellcolor{lightgray}3&\cellcolor{lightgray}  73.4/85.2 &\cellcolor{lightgray} 59.6/85.6 &\cellcolor{lightgray} 70.0/82.8&\cellcolor{lightgray}74.7/94.0 &\cellcolor{lightgray}67.1/85.4 \\
  5& 71.3/82.1  &  54.8/78.6 & 69.9/82.3 & 74.6/94.0 & 64.1/80.9 \\   
  \bottomrule
    \end{tabular}%
    }
  \label{tab:temporal}%
\end{table}%

\begin{table}[htbp]
   \small
  \centering
   \vspace{-0.2cm}
  \caption{Ablations for dynamic mask weight.}
   \setlength{\tabcolsep}{1pt}
   \scalebox{0.98}{
      \begin{tabular}{l|ccccc}
    \toprule
    Methods & Car & Pedestrian & Van & Cyclist & Mean \\
    \hline
    w/o $\beta$ & 71.5/84.4 & 58.1/85.7 & 71.9/83.8 & 64.4/89.7 & 65.6/85.0 \\
     \cellcolor{lightgray}with $\beta$ & \cellcolor{lightgray}73.4/85.2 & \cellcolor{lightgray} 59.6/85.6 &  \cellcolor{lightgray} 70.0/82.8& \cellcolor{lightgray} 74.7/94.0 & \cellcolor{lightgray} 67.1/85.4 \\
    LM & 73.8/85.3 & 56.4/83.9 & 72.6/84.6 & 72.7/93.0 & 66.1/84.8 \\
    \bottomrule
    \end{tabular}%
    }
  \label{tab:mask}%
\vspace{-0.3cm}
\end{table}%

In this section, we conduct extensive ablation studies on \nameo using the KITTI dataset under the category-unified setting, analyzing the impact of the model's core components and settings. Additional ablation results are provided in the supplementary material.\\
\noindent{\textbf{Model components.}} Table~\ref{tab:components} presents the ablation study of the components in \name. The experimental results indicate that when RECON is fully finetuned, the average metric of the model is limited. However, when we freeze the parameters of RECON and introduce our two-path adapter for transfer learning, the average metric improves significantly. Furthermore, by progressively incorporating our MoGE, learnable temporal token, and dynamic weighted mask, the performance continues to increase, demonstrating the effectiveness of each component.\\
\noindent{\textbf{Position of MoGE and Adapters.}} We analyze the impact of insertion positions for adapters and MoGE in Table~\ref{tab:position}. For adapters, we observe that performance improves with the number of layers they are added to; thus, we ultimately incorporate them into all layers. In contrast, for MoGE, adding it to all layers results in an obvious performance drop, which we attribute to overfitting caused by excessive parameters. The best performance is achieved when MoGE is applied to half of the layers, and this is our final setting. \\
\noindent{\textbf{Length of Temporal Propagation.}} We investigate the impact of the number of sampled frames for training temporal tokens in Table~\ref{tab:temporal}. Note that setting the number of propagation frames to 2 corresponds to not using temporal tokens. It is observed that using 3 frames yields the best performance. However, further increasing the sequence length does not lead to performance gains, suggesting that overly long search video clips impose a learning burden on the model. Therefore, it is important to choose an appropriate length for the search video clip.\\
\noindent{\textbf{Influence of Dynamic Mask Weighting.}} In Table~\ref{tab:mask}, we explore the effects of masks under different settings. Here, ``w/o $\beta$" indicates directly using the mask, ``with $\beta$" refers to our dynamic weighting mask, and ``LM" means initializing the mask as a learnable parameter. The results show that incorporating learnable components into the mask is beneficial. Moreover, multiplying the mask by an additional learnable $\beta$ achieves better performance compared to directly making the mask fully learnable. \\
\vspace{-.4cm}
\subsection{Visualization Analysis}
\begin{figure}
 
  \centering
  \includegraphics[width=1\linewidth]{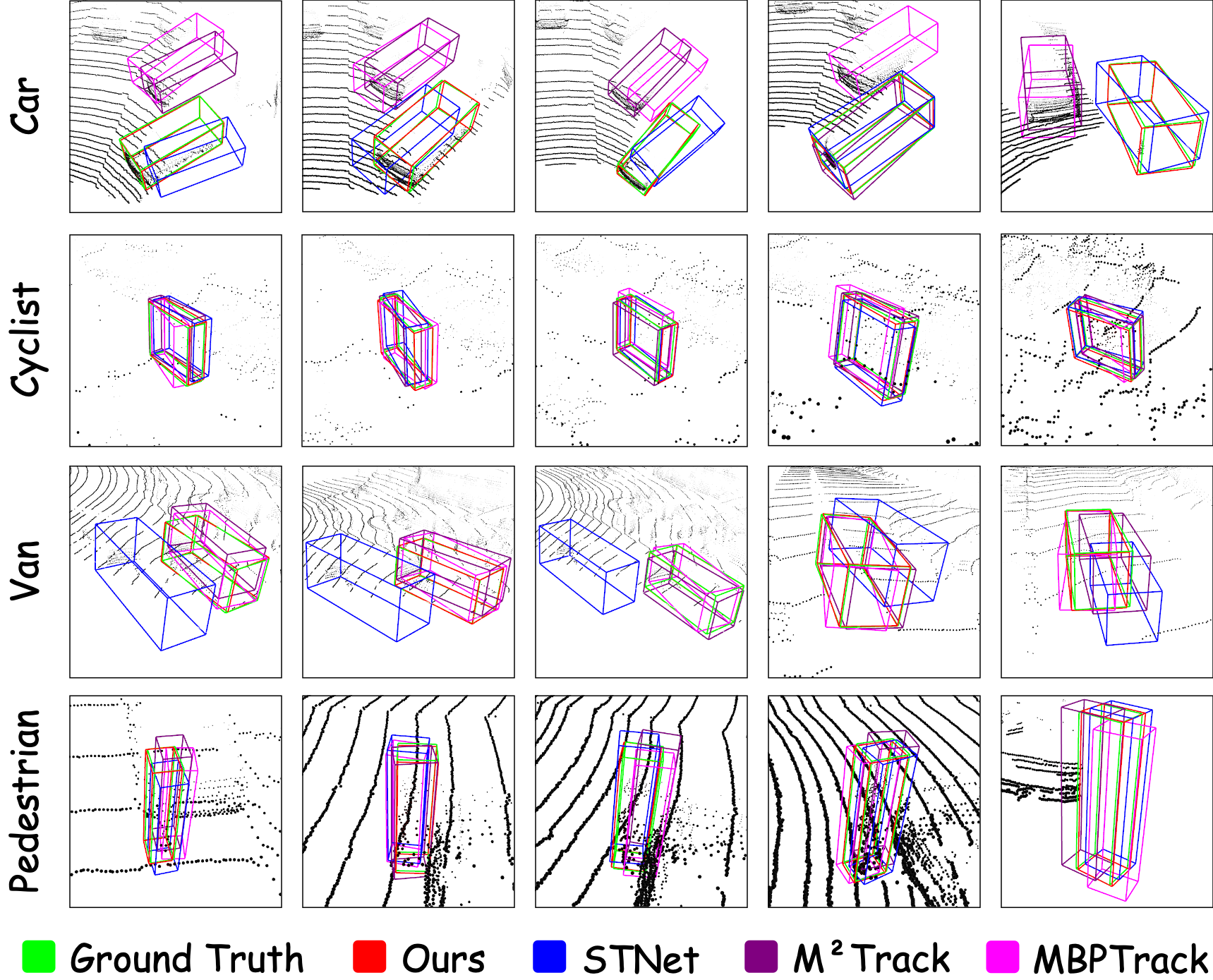} 
  \caption{\small \textbf{Qualitative Visualization.} }\label{fig:visualization}
  \vspace{-.6cm}
\end{figure}
In Fig.~\ref{fig:visualization}, we present the qualitative results on KITTI under category-unified settings. It is evident that even when dealing with scenarios where object surface points are extremely sparse, such as those involving Car and Van, our method still demonstrates significant advantages. Our \nameo closely aligns with the ground truth, whereas other methods exhibit noticeable deviations or lose track of the target. Additionally, in the tracking of Pedestrian and Cyclist, our method also shows higher accuracy and stability. Even in scenes with dense crowds or background interference, our method can continuously provide reliable target localization without being easily affected by surrounding environmental disturbances.
These qualitative results not only validate the superiority of our method in handling geometric disparities but also demonstrate its effectiveness and reliability in complex environments.
\section{Conclusion}
\vspace{-.1cm}
In this paper, we present \name, the first framework to bridge large-scale pretrained point cloud models with category-agnostic 3D SOT. By integrating parameter-efficient adapters and a geometry-aware expert architecture, \nameo effectively transfers geometric priors from pretraining while adaptively resolving cross-category disparities. The proposed temporal context optimization further enhances robustness to temporal changes and feature calibration. Extensive experiments demonstrate that \nameo achieves SOTA performance under category-unified settings while remaining competitive with category-specific methods, showcasing strong generalization across diverse real-world scenarios.

\section*{Acknowledgment}This work was supported by the National Natural Science Foundation of China under Grant No.~62403429, No.~62476247, No.~62402442, and the Zhejiang Provincial Natural Science Foundation of China under Grant No.~LQN25F030008, No.~LQ24F020038, and the Zhejiang Provincial ``Jianbing Lingyan + X'' Science and Technology Program No.~2025C01030.
{
    \small
    \bibliographystyle{ieeenat_fullname}
    \bibliography{main}
}

\end{document}